\documentclass[10pt,twocolumn,letterpaper]{article}

\usepackage{cvpr}
\usepackage{times}
\usepackage{epsfig}
\usepackage{graphicx}
\usepackage{amsmath}
\usepackage{amssymb}
\usepackage{algorithm}
\usepackage{algorithmicx}
\usepackage{algpseudocode}
\usepackage{subfigure}
\usepackage{color}
\usepackage{booktabs}
\usepackage{authblk}

\newtheorem{theorem}{Theorem}

\usepackage[pagebackref=true,breaklinks=true,letterpaper=true,colorlinks,bookmarks=false]{hyperref}

\cvprfinalcopy % *** Uncomment this line for the final submission

\def\cvprPaperID{****} % *** Enter the CVPR Paper ID here

\ifcvprfinal\fi
\begin{document}

%%%%%%%%% TITLE
\title{PDA: Progressive Data Augmentation for General Robustness \\of Deep Neural Networks}

\author[1]{Hang Yu\footnote[1]}
\author[1]{Aishan Liu\footnote[1]}
\author[1]{Xianglong Liu\footnote[2]}
\author[1]{Gengchao Li}
\author[2]{Ping Luo}
\author[3]{\\Ran Cheng}
\author[1]{Jichen Yang}
\author[1]{Chongzhi Zhang}
\affil[1]{State Key Laboratory of Software Development Environment, Beihang University}
\affil[2]{The University of Hong Kong}
\affil[3]{Southern University of Science and Technology}

\maketitle
\renewcommand{\thefootnote}{\fnsymbol{footnote}}
\footnotetext[1]{Contribute Equally.}
\footnotetext[2]{Corresponding author.}
%\thispagestyle{empty}

%%%%%%%%% ABSTRACT
\begin{abstract}
    Adversarial images are designed to mislead deep neural networks (DNNs), attracting great attention in recent years. Although several defense strategies achieved encouraging  robustness against adversarial samples, most of them fail to improve the robustness on common corruptions such as noise, blur, and weather/digital effects (\eg frost, pixelate).
    %\pluo{what corruptions?}.
    To address this problem, we propose a simple yet effective method, named Progressive Data Augmentation (PDA), which enables general robustness of DNNs by progressively injecting diverse adversarial noises during training. In other words, DNNs trained with PDA are able to obtain more robustness against both adversarial attacks as well as common corruptions than the recent state-of-the-art methods.
    We also find that PDA is more efficient than prior arts and able to prevent accuracy drop on clean samples without being attacked.
    %\pluo{Is this what you mean?}.
    Furthermore, we theoretically show that PDA can control the perturbation bound and guarantee better generalization ability than existing work.
    Extensive experiments on many benchmarks such as CIFAR-10, SVHN, and ImageNet demonstrate that PDA significantly outperforms its counterparts in various experimental setups.
    %promise more general robustness in practice: (1) on CIFAR-10, SVHN, ImageNet, it performs comprehensively well on adversarial noises compared with various augmentation methods, (2) it achieves state-of-the-art corruption robustness on the CIFAR-10-C and ImageNet-C benchmarks, (3) it enables deep models more evenly robustness in the frequency-based analysis. %Moreover, we also propose Mixed Test to evaluate model generalization ability more fairly.
\end{abstract}

%%%%%%%%% BODY TEXT
\section{Introduction}

Recent advanced deep neural networks (DNNs) achieve great successes in many fields including computer vision \cite{krizhevsky2012ImageNet}, natural language processing \cite{bahdanau2014neural}, and speech processing \cite{Hinton2012Deep}. Their performances are typically obtained by training with sufficient amount of `clean' data.
%
%===================Fig1================================
\begin{figure}[tp!]
%\vspace{-0.1in}
\centering
\vspace{-0.2in}
\subfigure[Some corrupted images in ImageNet-C]{
\includegraphics[width=0.95\linewidth]{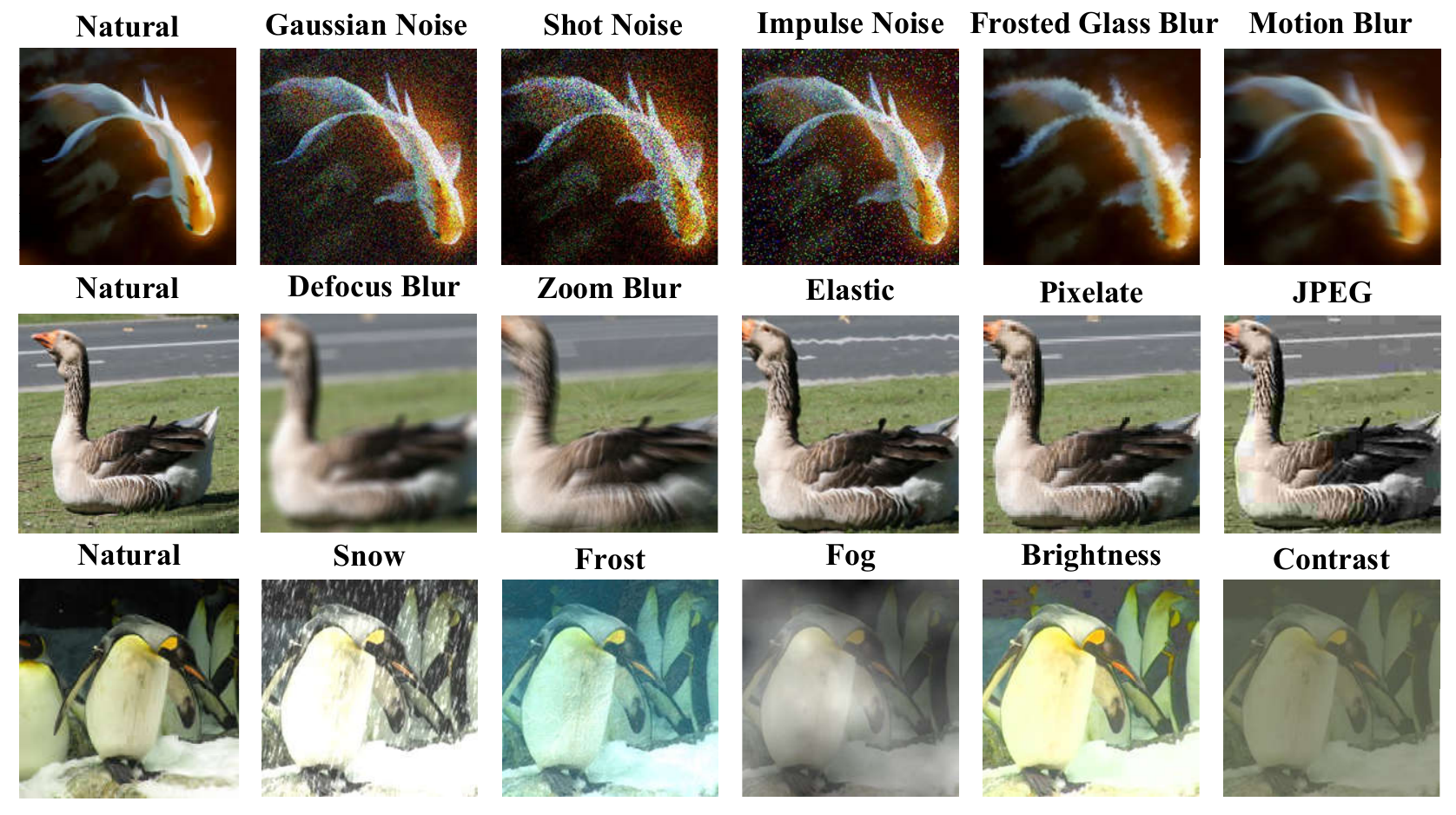}
}
%\hspace{-0.1in}
\subfigure[PGD \& GDA]{
\includegraphics[width=0.47\linewidth]{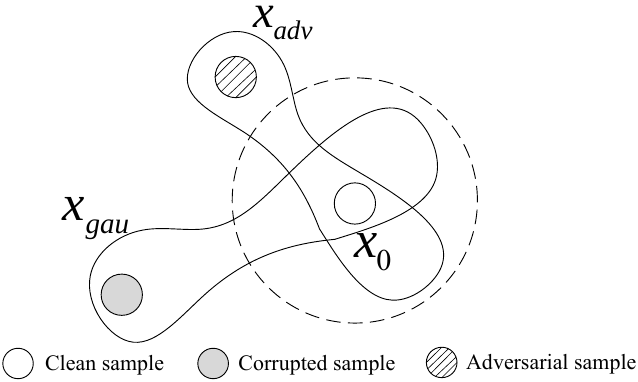}
}
%\hspace{0.05in}
\subfigure[PDA]{
\includegraphics[width=0.47\linewidth]{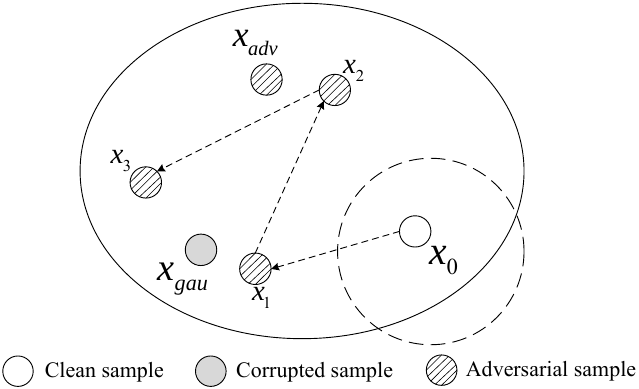}
}
\caption{An illustration of corrupted images (a) and the classification of models with PGD, GDA (b) and PDA (c). For both (b) and (c), the dashed curves represent the decision boundary of naturally trained models. Without covering the adversarial and corrupted samples promises better clean accuracy but almost no robustness. The solid curves denote the decision boundary after trained with augmented data. For PGD \& GDA (b), the decision space is ``stretched'' after augmentation (\eg $x_{adv}$ and $x_{gau}$) in some specific orientations, which causes the accuracy drop on clean samples. However, for our PDA (c), more diversified data are computed, aggregated and injected (\eg $x_{1}$, $x_{2}$, ... ) in various directions, which enlarges the decision space to cover adversarial, corrupted and clean samples simultaneously. The larger decision space of PDA promises the general robustness and prevents the drop of clean accuracy. In CIFAR-10, SVHN and ImageNet, PDA has comparable adversarial robustness and around 3\% above on clean accuracy when compared with PGD. In CIFAR-10-C and ImageNet-C, PDA has about 10\% above on mean Corruption Error (mCE) when compared with GDA.
%\\ As illustrated in Figure \ref{fig:fig1} (a), intuitively, the data partition $\hat{\mathcal{C}}_{i}$ for the $i$-th class could become moderately narrow after GDA and PGD, indicating that the partition space is ``stretched'' to cover more adversarial examples or Gaussian noises in some specific directions \cite{shaham2018understanding}. However, due to the limited capacity of data partition $\hat{\mathcal{C}}_{i}$, the model has poor generalization to benign examples. In contrast, as illustrated in Figure \ref{fig:fig1} (b), the capacity of data partition $\hat{\mathcal{C}}_{i}$ becomes larger after the model is trained with PDA. Hence, it is expected that PDA has better generalization to benign examples.
}
\label{fig:fig1}
\vspace{-0.1in}
\end{figure}
%=======================================================
However, in real-world scenario, the training and test data may contain many types of noises.
% often impractical to ,
For example, adversarial images (\eg adversarial sample generated by injecting small perturbation in an input image of DNN) and image corruptions \cite{hendrycks2018benchmarking} such as noise, blur, and weather/digital effects (\eg frost, pixelate) have been proven to threaten performance of DNNs \cite{Liu2019Perceptual,kurakin2016adversarial}, making them un-robust when deploying in scenarios that demand high security. This work intends to well understand and improve robustness of DNNs when significant adversarial samples and corruptions are presented.
%, which is an essential property especially for safety critical applications
In the recent years, many efforts have been devoted to improve robustness of DNNs %to the adversarial noises (or adversarial examples), maliciously constructed imperceptible perturbations that fool deep learning models,
in terms of \emph{attack} \cite{goodfellow6572explaining,athalye2018obfuscated} and \emph{defense} \cite{xie2018mitigating,alexey2017adversarialmachine,madry2017towards}. The attack approaches aim to inject noises or adversarial samples to confuse a DNN, while the defense approaches aim at preventing DNNs from being attacked. The existing work can be generally categorized into two streams.
In the first stream, the defense methods
attempt to build DNNs with adversarial robustness by gradient masking. Some of the defenses make classifier learn to break gradient descent, others are designed to purposely cause gradient masking as obfuscated gradients. %\pluo{what are those?}. %\pluo{Rewrite--supplying adversaries with non-computable gradients}.
Although they obtained certain stabilization for DNNs when adversarial samples are presented,
%encountering ,
for those methods designed to obfuscate gradients,
they still could be easily circumvented by constructing function to approximate the non-differentiable layer on the backward pass, or by approximating the expectation with samples over the transformation to correctly compute the gradient \cite{athalye2018obfuscated}. %\pluo{what are those?}.
%\pluo{reviewer don't understand without background knowledge}.
The recently proposed \emph{adversarial training} appears a strong defense algorithm by using adversarial data augmentation, showing strong capability to train robust DNNs against adversarial samples.
However, training on one-step adversarial examples does not confer robustness to iterative white-box attack \cite{kurakin2016adversarial} at ImageNet scale. It is possible that much larger models are necessary to achieve robustness to such a large class of inputs.
%
%they still failed to defence stronger iterative attacks\pluo{why?}.

In the second stream,
%Besides the progress in robustness to adversarial examples,
recent studies have paid attention to improve model robustness against common corruptions \cite{zheng2016improving,sun2018feature}. For example, corruptions like Gaussian blur and noise would greatly impede performance of DNNs as shown in \cite{dodge2017study,hendrycks2018benchmarking}.
%
%In real-world scenarios,  are more likely to be witnessed by practical deep learning systems. \cite{dodge2017study} found that deep learning models behave distinctly subhuman to input images with Gaussian noises. Likewise, deep learning models show weak performance on various corruptions including blur, pixelation and other types of noises \cite{hendrycks2018benchmarking}.
%
Although the above work achieved certain progress, they mainly focused on defending either corruptions \cite{zheng2016improving,sun2018feature} or adversarial samples \cite{alexey2017adversarialmachine,madry2017towards}, but seldom work have devoted to defend both of them.
%
%Though the existence of noises especially in real-world environment has drawn intense concerns about the robustness of deep models, the literature mainly focused on either corruptions \cite{zheng2016improving,sun2018feature} or adversarial noises \cite{alexey2017adversarialmachine,madry2017towards}. A very few studies have investigated the problem from the view of generalized noise robustness.
For example, only a few pioneering work \cite{fawzi2016robustness,fawzi2018adversarial,ford2019adversarial} try to establish relationship between adversarial samples and random noises (\eg additive Gaussian noise), leading to more general robustness of DNNs.
%
%\cite{fawzi2016robustness} studied the robustness of classifiers from adversarial examples to random noises and tried to build a robust model from the view of curvature constraints. \cite{fawzi2018adversarial} drew the relationship between in-distribution robustness and unconstrained robustness. More recently, \cite{ford2019adversarial} found that adversarial robustness is closely related to robustness under certain kinds of distributional shifts, i.e., additive Gaussian noise. Despite the promising progress, there still
It remains an open question to build strong DNNs that are able to against both adversarial samples and image corruptions by using a unified framework.
%
%that need to be answered for deeper understanding of model robustness: \emph{Could we build strong models against both adversarial noises and corruption as well?}

To understand DNNs' robustness and bridge the above gap, this work devises a novel algorithm, named \emph{Progressive Data Augmentation (PDA)}, whcih is a unified framework to improve robustness of DNNs against both adversarial samples and corruptions including noises, blur, weather and digital effects.
%
%more general augmentation named \emph{Progressive Data Augmentation} (PDA) to improve both adversarial and corruption robustness.
PDA is significantly different from prior arts that achieved robustness by using large-scale additional training data \cite{schmidt2018adversarially,sun2019towards}. Instead of simply augmenting data, PDA explores an alternative perspective by progressively aggregating, augmenting, and injecting adversarial noises during adversarial training.
%the diversified adversarial noises are aggregated, augmented and injected progressively, which are
We prove that PDA is beneficial to improve robustness against both adversarial samples and corruptions.
For example, we show that
%We also theoretically prove that
PDA is able to satisfy the perturbation bound and guarantee better generalization ability.
Furthermore, extensive experiments on multiple datasets including CIFAR-10 \cite{krizhevsky2009learning}, SVHN \cite{netzer2011reading}, ImageNet \cite{deng2009imagenet}, CIFAR-10-C \cite{hendrycks2018benchmarking} and ImageNet-C \cite{hendrycks2018benchmarking} demonstrate that PDA outperforms existing approaches in various experimental settings.
% which further indicate that PDA shows comprehensively excellent results on both adversarial and corruption noise compared to various augmentation methods.

\subsection{Related Work}
\noindent \textbf{Adversarial Robustness.} Many adversarial training based methods have been proposed to improve model robustness against adversarial examples.
%
%\cite{madry2017towards} adversarially trained moderately the most robust model with the PGD augmented adversarial examples, whereas it consumes too much time and fails to generalize well on clean examples.
\cite{madry2017towards} trained robust model with the PGD augmented adversarial samples, whereas it consumes too much time and fails to generalize well on clean examples.
Meanwhile, \cite{sinha2017certifiable} proposed an provable adversarial training algorithm with a surrogate loss, but it is only confined to the small dataset, \eg MNIST.
Besides, \cite{sankaranarayanan2018regularizing} calculated and added adversarial noises to each hidden layer during training to tackle the overfitting problem.
Moreover, \cite{shafahi2019adversarial} and \cite{zhang2019you} both brought efficient strategies on the conventional adversarial training algorithm to achieve adversarial robustness without much time cost.
However, seldom work achieved the robustness in more general situations (\eg corruptions), and the adversarial training based method still
%
%fails to handle more aggressive iterative attacks and
remains questionable to various corrupted samples.

\noindent \textbf{Corruption Robustness.} When it comes to training the robust model towards corrupted samples, rare effective approaches have been proposed.
\cite{zheng2016improving} utilized stability training to stabilize model behavior against small input distortions, and \cite{sun2018feature} employed feature quantization techniques to deal with distorted images such as motion blur, salt and pepper.
\cite{metz2019using} drew insights from meta-learning which uses a learned optimizer to build a robust model against input noises (\eg translations).
\cite{ford2019adversarial} found the positive correlation between adversarial and corruption robustness. However, the Gaussian data augmentation they used only has limited improvement on adversarial robustness.
More recently, \cite{lopes2019improving} devised patch Gaussian augmentation method to improve the corruption robustness with little side effects on clean accuracy,
yet the performance on adversarial robustness was not provided in this work.

\noindent \textbf{Data Complexity and Diversity.} Prior studies \cite{schmidt2018adversarially,sun2019towards} have shown that the sample complexity plays a critical role in training a robust deep model.
\cite{schmidt2018adversarially} concluded that the sample complexity of robust learning can be significantly larger than that of standard learning under adversarial robustness situation.
\cite{charles2019convergence} believed that adversarial training may need exponentially more iterations to obtain large margins compared to standard training.
Moreover, \cite{zhai2019adversarially} theoretically and empirically showed that with more unlabeled data, the model can be learned with better adversarial robust generalization.

\section{Background and Notation}
In this section, we provide the definitions of the adversarial robustness and corruption robustness, taking the widely studied and used convolutional neural networks in image classification as the basic deep models.
%set $\mathcal{S}=\{(x_{i},y_{i})\}_{1}^{n}$ with a feature vector $x$ $\in$ $\mathcal{X}$ and a label $y$ $\in$ $\mathcal{Y}$, or $\mathbb{R}^{D}\rightarrow \{1,...,K\}$ , where $D$ is the dimension of input examples, $\{1,...,K\}$ is the corresponding labels of output.
Given a training sample $x$ $\in$ $\mathcal{X}$ and label $y$ $\in$ $\mathcal{Y}$, the pair of $(x,y)$ from distribution $D$, the deep supervised learning model tries to learn a function $f$: $\mathcal{X}$ $\rightarrow$ $\mathcal{Y}$ as the mapping from input examples to output labels. Specifically, we use log-loss in the image classification problem: $$\ell(\theta;x,y)=-\sum_{i=1}^{m}y_{i}\log(\frac{e^{f(\theta;x,y_{i})}}{\sum_{j=1}^{m}e^{f(\theta;x,y_{j})}}),$$
where $\theta$ denotes the parameter of a given neural network,
and $f(\theta;x,y_{i})$ is the score given to the pair $(x,y_{i})$ by a function $f$. %\rcheng{what does a 'position' mean?}
%More precisely, $f_{S}(\theta;x,y)$ represents the prediction results of input $(x,y)$ after the model is trained on training set $\mathcal{S}$.

Specifically, \textbf{adversarial robustness} is defined as $\min_{\|\delta\|_{p}\leq\varepsilon} \mathbb{P}_{(x,y)\sim D}(f(x+\delta)=y)$, i.e.,  the probability of right classification when a random sample from distribution $D$ is perturbed by the small adversarial noise $\delta$ controlled by $\varepsilon$; and \textbf{corruption robustness} is defined as $\mathbb{P}_{(x,y)\sim D}(f(c(x)=y))$, i.e., the probability of right classification when a sample from distribution $D$ is corrupted by the transformation function $c(\cdot)$ \cite{hendrycks2018benchmarking}.

Generally, the adversarial attack by generating adversarial examples can be defined as:
$$x'=\mathop{\arg\max}\limits_{\|x'-x\|_{p}\leq\varepsilon}\ell(\theta;x,y),$$
where $\varepsilon$ represents the perturbation magnitude. The optimization process could be approximately solved with the first-order expansion \cite{shaham2018understanding}:
$$x'=\mathop{\arg\max}\limits_{\|x'-x\|_{p}\leq\varepsilon}\nabla_{x}(\ell(\theta;x,y))^{T}(x'-x).$$
In the case p = $\infty$, we obtain $x' = x + \varepsilon$sign$ (\nabla_{x}\ell(\theta;x,y))$, which is the famous attack Fast Gradient Sign Method (\textbf{FGSM} \cite{goodfellow2014explaining}). The iterative version of this generating process are proposed by \cite{kurakin2016adversarial} and \cite{madry2017towards}, which obtains stronger adversarial examples.

To enhance adversarial robustness, the \textbf{adversarial training} can be performed by solving:
$$\min_{\theta}\mathbb{E}_{(x,y)\sim D}\max_{\|\delta\|_{p}\leq\varepsilon}\ell(\theta;x+\delta,y),$$
where the process can be regarded as the augmentation of $x$ with the adversarial example $x + \delta$ in the normal training. %\rcheng{why? any evidence or reference?}
%\rcheng{what do you mean by 'attack'? up to now, there is no definition of 'attack'. Something seems missing here to bridge the gap between the contexts.}
Among various attack methods, the projected gradient descent \textbf{(PGD)} \cite{madry2017towards} is commonly used in the adversarial training as PGD-AT to achieve excellent adversarial robustness. However, the searching process of $\delta$ in PGD requires massive computation cost, which motivates us to propose a efficient method to achieve comparable adversarial robustness.  %\rcheng{so what? is it related to the motivation of the work?}

A common idea for enhancing the corruption robustness is to add Gaussian noise in the training process:
$$\min_{\theta}\mathbb{E}_{(x,y)\sim D}\ell(\theta;x+\mathcal{N}(x;\sigma^{2}I),y),$$
where the training process could be regarded as the Gaussian data augmentation \textbf{(GDA)} \cite{ford2019adversarial}. Despite of its simple implementation, GDA only has limited performance against adversarial examples and other corruptions beyond noise.

\section{Progressive Data Augmentation (PDA)}

From the data complexity point of view \cite{schmidt2018adversarially,sun2019towards,zhai2019adversarially}, %\rcheng{why?},
we introduce an adversarial-based method called \textit{Progressive Data Augmentation (PDA)}. During multiple iterations within each training step, PDA adds adversarial noises progressively with different magnitudes into training data. Different from PGD or GDA, our progressive data augmentation significantly increases the data diversity with more flexible perturbations during each training step. Therefore, the models trained with PDA are  expected to be generally more robust against more types of noises, including both adversarial examples and corruptions. Moreover, the process of generating augmented noises could be significantly faster than PGD-based method as it merely utilizes the training gradients with little extra computation cost. Therefore, in comparison with the searching process of PGD-based methods, PDA is more efficient in generating diversified noises.

%%%%%%%%%%%%%%%%%%%%%%%%%%%%%%%%%%%%%%%%%%%%%%%%%%%%%%%%%%%%%%%%%%%%%%

%%%%%%%%%%%%%%%%%%%%%%%%%%%%%%%%%%%%%%%%%%%%%%%%%%%%%%%%%%%%%%%%%%%%%
%For some $x_{0}$ in training set $S$, assuming the function $f_{S}(\theta;x_{0},y)$ of trained model should classify $x_{0}$ into some partition $\hat{C}_{i}$. If there are two different class $i,j$ for single-label classification problem, then we have $\hat{C}_{i} \cap \hat{C}_{j}=\emptyset$.
%Inspired by the theoretical connections between adversarial and corruption robustness, we propose our \emph{Progressive Data Augmentation} (PDA) strategy that adds adversarial noises with huge complexity during training to shrink $\mathcal{R}_{adv}$ significantly, which leads to the larger increase of $(\mathcal{C}_{i}^{T})^{\varepsilon} \backslash \mathcal{C}_{i}^{T}$ with the fixed $\mathcal{C}_{i}$ resulting in the decrease of $\mathcal{R}_{cor}$. Thus, both adversarial and corruption robustness are guaranteed.

The surrogate loss was first proposed in \cite{sinha2017certifiable}, by adding the Lagrangian constraint to the original loss function  to approximate the worst-case perturbation. Analogously, we can update the model parameter $\theta$ and $\delta$ by minimizing the following surrogate loss:
\begin{equation}
 \ell_{\lambda}(\theta;x,y) = \min_{\theta}\sup_{\delta}[\ell(\theta;x,y,\delta)-\frac{\lambda}{2}\|\delta\|_{2}^{2}],
\end{equation}
where $\delta$ represents the added perturbation which helps approximate the objective robust distribution.

Specifically, given a pair of training samples $(x,y)$, we assume $\delta$ as the perturbation to be updated with gradient ascent. %\rcheng{do you mean gradient descent?}
Since data complexity contributes significantly to adversarially robust model, we propose a progressive iteration process to introduce diversified adversarial examples. Given the fact that $k\|\nabla_{\delta}\ell(\theta;x,y,\delta)\|_{2}$=$\|k\nabla_{\delta}\ell(\theta;x,y,\delta)\|_{2}$ and $ \nabla_{\delta}\ell_{\lambda}(\theta;x,y) = \nabla_{\delta}\ell(\theta;x,y,\delta)-\lambda\|\delta\|_{2}$, we have
 \begin{equation}\label{eq:6}
 \delta \gets \delta + \frac{\varepsilon}{\|k\nabla_{\delta}\ell(\theta;x,y,\delta)\|_{2}} \cdot \nabla_{\delta}\ell_{\lambda}(\theta;x,y),
\end{equation}

Thus, the perturbation $\delta$ could be updated in every step $j$ during the progressive process as:
 \begin{equation}\label{eq:5}
 \delta^{j} \gets (1-\lambda)\delta^{j-1} + \frac{\varepsilon}{k} \cdot \frac{\nabla_{\delta^{j-1}}\ell(\theta;x^{j-1},y,\delta^{j-1})}{\|\nabla_{\delta^{j-1}}\ell(\theta;x^{j-1},y,\delta^{j-1})\|_{2}},
\end{equation}
where $\lambda$ corresponds to the $l_{2}$ regularization factor and the magnitude $\varepsilon$ is normalized by the number of steps $k$, such that the overall magnitude of update equals $\varepsilon$. $\lambda$ here controls perturbation decay with the increase of progressive iterations as well as the contribution of the former perturbation $\delta^{j-1}$. Then we augment the current training sample  as $x^{j} \gets x^{j-1} + \delta^{j}$, and update model parameter with the augmented $x^{j}$, progressively.

%where $\lambda$=$\beta\|k\nabla_{\delta^{*}}\ell(\theta;s)\|/\varepsilon$,

In the training epoch $t$ we diversified the overall perturbation magnitude $\varepsilon^{t}$ as $\{0,\frac{\varepsilon}{3},\frac{\varepsilon}{2},\varepsilon,\frac{\varepsilon}{2},\frac{\varepsilon}{3},0\}$. For the first half of training process, the ascent of magnitude is similar to the curriculum learning \cite{bengio2009curriculum}, which makes the model easier to learn. For the second half of training epochs, the descent of magnitude could help fix the decision boundary and make it smoother, which is beneficial to the classification of benign samples.

%During each perturbation computation step via gradient ascent, another batch of adversarial examples is obtained as $x \gets x + \delta^{*}$, thus another one-step gradient descent is executed to promote the model parameters update.
%\begin{equation}
%$x \gets x + \delta^{*}$
%\end{equation}
%Obviously, $k\|\nabla_{\delta^{*}}\ell(\theta;s)\|$=$\|k\nabla_{\delta^{*}}\ell(\theta;s)\|$. With the surrogate loss optimizing, we have
% \begin{equation}\label{eq:6}
% \delta^{*} \gets \delta^{*} + \frac{\varepsilon}{\|k\nabla_{\delta^{*}}\ell(\theta;s,\delta^{*})\|} \cdot [\nabla_{\delta^{*}}\ell(\theta;s,\delta^{*})-\lambda\delta^{*}],
%\end{equation}
%where $\lambda$=$\beta\|k\nabla_{\delta^{*}}\ell(\theta;s)\|/\varepsilon$, according to equation \eqref{eq:5} and \eqref{eq:6}. In this way, Progressive Data Augmentation can be regarded as the optimization of robust surrogate loss in an iterative way. %\textcolor{red}{???the derivation ???}

%=========================Algorithm===============================
%\vspace{-0.15in}
\begin{algorithm}
    \caption{Progressive Data Augmentation}
    \label{algorithm:1}
    \begin{algorithmic}[1]
        \Require Training set $\{(x_{i},y_{i})\}_{i=1}^{n}$, Hyper-parameters $\lambda$, $\varepsilon$, $k$, learning rate $\eta$
        \Ensure Updated model parameters $\theta$
        \For{$t$ in $T$ epochs}
            \State $\varepsilon^{t} \in \{0,\frac{\varepsilon}{3},\frac{\varepsilon}{2},\varepsilon,\frac{\varepsilon}{2},\frac{\varepsilon}{3},0\}$ ,\quad for diversification
            \For{$j$ in $k$ iterative steps}
                \State $\delta^{j} \gets (1-\lambda)\delta^{j-1} + \frac{\varepsilon^{t}}{k} \cdot \frac{\nabla_{\delta^{j-1}}\ell(\theta;x^{j-1},y,\delta^{j-1})}{\|\nabla_{\delta^{j-1}}\ell(\theta;x^{j-1},y,\delta^{j-1})\|_{2}}$
                \State $x^{j} \gets x^{j-1} + \delta^{j}$
                \State $\theta \gets \theta - \eta \nabla_{\theta}\ell(\theta;x^{j},y,\delta^{j})$
            \EndFor
        \EndFor
    \end{algorithmic}
\end{algorithm}
%\vspace{-0.15in}
%=================================================================

%-------------------------------------------------------------------
%\section{Theoretical Analysis}
%\label{others}
%------------------------------------------------------------------

\section{Theoretical Analysis}

Now, we further analyse the possible perturbation bound with respect to robustness and the upper bound on the expected generalization error. For the loss function $\ell(\theta;x,y)$, obviously there exist upper bounds $M_{0}$ and $M_{1}$, such that $|\ell(\theta;x,y)|\leq M_{0}$ and $\|\nabla_{x}\ell(\theta;x,y)\|_{2} \leq M_{1}$.

%\begin{assumption}\label{Assumption 1}
Assuming that the loss function is smooth, there exist $L_{0}$ and $L_{1}$ such that $|\ell(\theta;x,y)-\ell(\theta;x',y')|\leq L_{0}\|x-x'\|_{2}$, $\|\nabla_{x}\ell(\theta;x,y)-\nabla_{x'}\ell(\theta;x',y')\|_{2} \leq L_{1}\|x-x'\|_{2}$, where $(x,y)$ and $(x',y')$ denote two pairs of samples.
%\end{assumption}
\paragraph{Perturbation Bound.}
We set $z = g(\theta;x)$ as the mapping from the input $x$ to the output of last hidden layer $z$. By defining the output set of $z$ as $\mathcal{Z}$ and rewriting $\ell(\theta;x,y)$ as $\ell(\theta;z,y)$ with the variable $z$, we define $z_{\varepsilon}^{*}$ as the $\varepsilon $- maximizer \cite{bonnans2013perturbation} of $\ell(\theta;z,y)$: %\rcheng{count how many 'We' you have used here}
$z_{\varepsilon}^{*} \in \varepsilon - arg\max_{z\in\mathcal{Z}}\{\ell(\theta;x,y)\}$. Given a pair of sample $(x_{0},y_{0})$, $z_{0}$ denotes the output of last hidden layer for sample $x_{0}$. Then we have:

\begin{theorem}\label{theorem:1}
In the intersection $U$ of the neighborhoods of $z_{0}$ and $z_{0} - (\nabla_{zz}\ell(\theta;z_{0};y_{0}))^{-1}\nabla_{z}\ell(\theta;z_{0},y_{0})$, we assume:
\begin{enumerate}
\item There exists $C$, s.t. $\forall z \in U$, $\|\nabla_{zz}\ell(\theta;z,y)\|_{2}\geq C$;
\item There exists $K$, s.t. $\forall z \in U$, $\|(\nabla_{z}\ell(\theta;z,y_{0})-\nabla_{z}\ell(\theta;z_{0},y_{0}))-\nabla_{zz}\ell(\theta;z_{0},y_{0})(z-z_{0})\|_{2} \leq K$.
\end{enumerate}
With the above assumptions, we have the \textit{perturbation bound} as follows:
\begin{small}
$$\|z_{\varepsilon}^{*}-z_{0}\|_{2} \leq \frac{K}{C} + \sqrt{\frac{\varepsilon}{C}} + \|(\nabla_{zz}\ell(\theta;z_{0},y_{0}))^{-1} \nabla_{z}\ell(\theta;z_{0},y_{0})\|_{2} $$
\end{small}
%where $Sec = \|(\nabla_{zz}\ell(\theta;z_{0},y_{0}))^{-1} \nabla_{z}\ell(\theta;z_{0},y_{0})\|_{2}$.
\end{theorem}
The detailed proof of Theorem \ref{theorem:1} can be found in the appendix. With the definition of $\varepsilon $- maximizer and the above assumptions, we are able to find the proper space in some neighborhood to satisfy the conditions, and thus compute the output range of the last hidden layer. Therefore, with Theorem \ref{theorem:1}, we could estimate the impact of adversarial examples and the potential risk with the distance between the worst-case perturbed one and the normal one. %\rcheng{how? The reason is unclear. If there is no solid reason, please revise or delete this conclusion}

\paragraph{Generalization Ability Bound.}
%\begin{lemma}\label{Lemma 1}
%([\cite{xu2012robustness},Theorem 14]).
%For the set $\mathcal{Z}$ and $l_{2}$ metric $\|\cdot\|$, if $\ell(\theta;z)$ satisfies $L_{0}$-Lipschitzian, then for augmented training set $\mathcal{S}$ and $\forall \gamma > 0$,  $f_{\mathcal{S}}(\theta;\cdot)$ is $(N(\gamma/2,\mathcal{Z},\|\cdot\|),\gamma L_{0})$-robust. That is to say, $\mathcal{Z}$ can be partitioned into $K=N(\gamma/2,\mathcal{Z},\|\cdot\|)$ disjoint sets $\{\mathcal{C}_{i}\}_{i=1}^{K}$, such that $\forall s \in \mathcal{S}$,
%$$ s,z \in \mathcal{C}_{i} \Rightarrow |\ell(\theta;s)-\ell(\theta;z)| \leq \gamma L_{0}. $$
%end{lemma}
%\begin{theorem}\label{Theorem 1}
Specifically, we denote $N(\gamma,\mathcal{X},\|\cdot\|_{p})$ as the covering number of $\mathcal{X}$ using $\gamma$-balls for $\|\cdot\|_{p}$. Inspired by the Theorem 3 in \cite{xu2012robustness}, we have:

\begin{theorem}\label{theorem:2}
Given augmented training set $\mathcal{X}$, for any probability $p>0$, there is at least probability $1-p$ such that surrogate loss satisfies
\begin{equation*}
\begin{split}
  \ell_{\lambda}&(\theta;x,y) \leq \frac{1}{n}\sum_{i=1}^{n}\hat{\ell}(\theta;x,y)+ \gamma (L_{0}+\frac{2M_{1}L_{1}+1}{\lambda}) \\
  &+ M_{0}(\sqrt{\frac{2N(\gamma/2,\mathcal{X},\|\cdot\|_{2}) \ln 2-2\ln p}{n}})+\frac{M_{1}^{2}}{\lambda-L_{1}},
\end{split}
\end{equation*}
where $n=|\mathcal{X}|$ denotes the volume of training set and $\hat{\ell}$ denotes the empirical training loss.
\end{theorem}

Furthermore, the theoretical analyses of generalization error upper bound for the surrogate loss of PDA can be found in the appendix. Let us denote the Lipschitz constant of $\ell_{\lambda}(\cdot)$ as $L_{\lambda} = L_{0}+\frac{2M_{1}L_{1}+1}{\lambda}$, then the difference between losses could be constrained in $\gamma L_{\lambda} $ with similar input examples, and there is a balance between the covering number $N(\gamma/2,\mathcal{X},\|\cdot\|_{2})$ and $\gamma L_{\lambda}$. If $\gamma$ increases, then the former covering number decreases and $\gamma L_{\lambda}$ increases at the same time, which promises a reasonable generalization bound with the robust property. Therefore, according to  Theorem \ref{theorem:2}, PDA could build robust model with proved upper bound on the generalization error.

\section{Experiments}
In this section, we will evaluate our \textit{Progressive Data Augmentation} and other strategies on the image classification task against adversarial and corrupted examples.

We adopt the popular \textbf{CIFAR-10} \cite{krizhevsky2009learning}, \textbf{SVHN} \cite{netzer2011reading} and \textbf{ImageNet} \cite{deng2009imagenet} as the evaluation datasets of adversarial robustness. As for the deep models, we choose the widely-used models, VGG16 \cite{simonyan2014very} and Wide ResNet34 for CIFAR-10, standard ResNet18 \cite{he2016deep} for SVHN and AlexNet \cite{krizhevsky2012ImageNet} for ImageNet respectively. For simplicity, we only choose 200 out of the 1000 classes in ILSVRC-2012 with 100K images for training set and 10k for validation set. All experiments are taken on the single NVIDIA Tesla V100 GPU.

For the experiments of corruption robustness, we evaluate our proposed method on the benchmark \cite{hendrycks2018benchmarking} \textbf{CIFAR-10-C} and \textbf{ImageNet-C}. CIFAR-10-C and ImageNet-C are the first datasets for benchmarking model robustness against different common corruptions with different severity levels. They are created from the test set of CIFAR-10 and validation set of ImageNet. There are 15 kinds of corruptions in the dataset we have used, comprising the sets of Noise, Blur, Weather and Digital. Specifically, Noise contains \{Gaussian, Shot, Impulse\}, Blur contains \{Defocus, Glass, Motion, Zoom\}, Weather contains \{Snow, Frost, Fog, Bright\}, Digital contains \{Contrast, Elastic, Pixel, JPEG\}. Each corruption has 5 severities.

For the experiments of adversarial robustness in CIFAR-10 and SVHN, we evaluate the augmented model with Projected Gradient Descent (\textbf{PGD} \cite{madry2017towards}) on $l_{inf}$ norm and the \textbf{C\&W} attack \cite{carlini2017towards} on $l_{2}$ norm. The \textit{PGD-$k$} attack means the PGD attack with iteration steps $k$. For PGD attack, we set the overall magnitude of perturbation $\varepsilon$ as 4/255, 8/255 and 12/255 (4,8,12 for short) respectively, and set the iteration steps $k$=20. For C\&W attack, we set the $l_{2}$ perturbation size $c$=500, which is similar to \cite{zhang2019you}. For ImageNet, we evaluate the models with PGD-20 attack and BPDA \cite{athalye2018obfuscated} attack, where the perturbation magnitude $\varepsilon$=8/255 (8 for short).

For CIFAR-10, SVHN and ImageNet, we evaluate the performance of model with our progressive data augmentation (\textbf{PDA}), compared with naturally trained model (\textbf{Naturally Trained} or \textbf{Natural} for short), model with projected gradient descent (\textbf{PGD}) and model with Gaussian data augmentation (\textbf{GDA} \cite{ford2019adversarial}). We use \textit{PGD-$k$-$\alpha$} to represent the model augmented by PGD with iteration steps $k$ and attack step size $\alpha$, where we set $k$=5 or 10 and $\alpha$=1/255 (1 for short). We use \textit{GDA-$\sigma$} to represent the model augmented by Gaussian noise $\mathcal{N}(\cdot;\sigma^{2}I)$, where we set $\sigma$=0.1 or 0.4 in the training process. For our method, we use \textit{PDA-$k$-$\varepsilon$} to represent the model augmented by more varied noise with perturbation magnitude $\varepsilon$ and iteration times $k$.

%Following \cite{alexey2017adversarialmachine,athalye2018obfuscated,carlini2019evaluating}, we construct adversarial examples with the strongest attacks in different norms from both gradient-based attack and optimization-based attack including: Projected Gradient Descent (PGD) \cite{madry2017towards} attack and C\&W attack \cite{carlini2017towards}. For CIFAR-10 and SVHN, we set the $l_{inf}$-perturbation size $\varepsilon$ of PGD as 2/255 and 4/255, and
%For the compared methods, different data augmentation methods are employed including PGD-AT and GDA. %More experimental settings are listed in the supplementary material.

%Following \cite{hendrycks2018benchmarking}, we constructed corrupted datasets of CIFAR-10 and SVHN consisting of various types corruption, e.g., noise, blur,, weather, digital, etc.

\subsection{Evaluation Criteria}

For adversarial robustness, we evaluate the model robustness with top-1 classification accuracy on adversarial datasets. For corruption robustness, \cite{hendrycks2018benchmarking} proposed some different metrics to score the performance of a classifier, which could comprehensively evaluate a classifier's robustness to corruption.

Given a trained classifier $f$ which has not been trained on corruption set, we compute the top-1 error rate on clean images as $E^{f}_{clean}$. Then we test the classifier on each corruption $c$ at each level of severity $s$, denoted as $E^{f}_{s,c}$, and divide by the error rate of baseline $E^{base}_{s,c}$ for adjustment. Finally, the mean Corruption Error (\textbf{mCE}) is computed as:
\begin{small}
$$CE_c^f = \frac{\sum_{s=1}^5 E^f_{s,c}}{\sum_{s=1}^5 E^{base}_{s,c}},$$
\end{small}
which means the average of 15 different Corruption Errors (CE). Moreover, the amount that the classifier declines on corrupted inputs can be measured with \textbf{Relative mCE.} (RmCE for short). If a classifier withstands the most corruption, the gap between mCE and the clean error rate is minuscule. Thus, Relative mCE is calculated as:
\begin{small}
$$RmCE_c^f = \frac{\sum_{s=1}^5 E^f_{s,c} - E^f_{clean}}{\sum_{s=1}^5 E^{base}_{s,c} - E^{base}_{clean}}.$$
\end{small}
%where Relative mCE is the average score of them.
%Here we choose the baseline as the Naturally Trained model.

%\textbf{Experiment settings.} For CIFAR-10 we set the volume of the sample set as 10000, and for SVHN we set the volume as 26032, which are the same as the official dataset. We use SGD with momentum as an optimizer for ERM baseline, the compared methods and our method.
%The training details of the compared methods and PDA are reported in the supplementary material.

%In corruption set, we use all given severities by \cite{hendrycks2018benchmarking} to combine into whole set for different types,
%------------------------------------------------------------
\subsection{Adversarial Robustness}
%To comprehensively evaluate model robustness, we conduct the experiment on clean, adversarial and corrupted datasets, respectively. and we compare our PDA with Naturally Trained, PGD-AT and GDA.
\paragraph{CIFAR-10} We conduct the experiments of adversarial robustness on CIFAR-10 to demonstrate the effectiveness of our proposed PDA. Table \ref{tab:tab1} shows that PDA achieves state-of-the-art results on both of these adversarial attack benchmarks. Besides the performance against adversarial attacks, we could notice that diversified noises augmented in training data remarkably improve the model performance on clean images, as illustrated in Figure \ref{fig:fig1}. With the increase of perturbation magnitude for these augmentation strategies, there is a trade-off between clean accuracy and adversarial performance, indicating that our PDA is more suitable in the general case. We further compare the  computation cost of PDA with PGD on VGG16 and CIFAR-10. As indicated by the results in Table \ref{tab:tab8}, PDA is computationally more efficient than PGD in the same training epochs.

%---------------------------------Tab 1----------------------
\begin{table}[h!]
\caption{The adversarial robustness of naturally trained model and models with PGD, GDA and PDA using VGG16 on CIFAR-10. Following the guidelines from \cite{carlini2019evaluating}, models are evaluated with the gradient-based attack (PGD) in $l_{inf}$ norm and optimization-based attack (CW) in $l_{2}$ norm. PDA performs comprehensively well on adversarial attack as well as clean images.}
\label{tab:tab1}
\begin{small}
\begin{tabular}{lllll}
\hline
\multicolumn{1}{l}{VGG16} & \multicolumn{1}{c}{Clean} & \multicolumn{2}{c}{\begin{tabular}[c]{@{}c@{}}PGD-20 Attack \\ ($\varepsilon$=8, 12)\end{tabular}} & \multicolumn{1}{c}{\begin{tabular}[c]{@{}c@{}}CW Attack\\ (c=500)\end{tabular}} \\ \hline
Natural                   & \textbf{92.52\%}          & 0.17\%                                     & 0.00\%                                     & 6.72\%                                                                          \\
PGD-5-1 \cite{madry2017towards}                  & 87.60\%                   & 44.78\%                                    & 28.03\%                                    & 45.01\%                                                                         \\
GDA-0.1 \cite{ford2019adversarial}                  & 89.14\%                   & 11.61\%                                    & 2.05\%                                     & 55.15\%                                                                         \\ \hline
PDA-3-1.0                 & 90.94\%                   & 41.06\%                                    & 25.08\%                                    & \textbf{67.17\%}                                                                \\
PDA-3-2.0                 & 89.56\%                   & \textbf{44.87\%}                           & \textbf{33.91\%}                           & 55.19\%                                                                         \\ \hline
\end{tabular}
\end{small}
\end{table}
%-------------------------------------------------------------
%\vspace{-0.1in}
%---------------------------------Tab 8----------------------
\begin{table}[h!]
\caption{The time consumption of training process with VGG16 on CIFAR-10. PDA achieves 1/6 to 1/5 computation cost of PGD.}
\label{tab:tab8}
\begin{small}
\begin{center}
\vspace{-0.1in}
\begin{tabular}{lcc}
\hline
Time cost (mins) for 80 epochs & PGD-5-1 \cite{madry2017towards} & PDA-3-2.0       \\ \hline
VGG16                          & 174.424 & \textbf{30.572} \\ \hline
\end{tabular}
\vspace{-0.1in}
\end{center}
\end{small}
\end{table}
%------------------------------------------------------------

Similarly, we compare our PDA method with other two current methods `Free-m' \cite{shafahi2019adversarial} and `YOPO' \cite{zhang2019you}. YOPO compared the while-box PGD-20 attack performance and time consumption with Free-m using Wide ResNet34 on CIFAR-10.
%{\rcheng{???}}
As shown in Table \ref{tab:tab9}, PDA achieves comparable performance on adversarial attack with less time cost. Compared with other augmented strategies, PDA still has better accuracy on clean images due to more varied data in augmentation.

%---------------------------------Tab 9----------------------
\begin{table}[h!]
\caption{The adversarial robustness and time cost of Wide ResNet34 using different strategies on CIFAR-10. PDA achieves comparable adversarial performance and computation cost as well as better clean accuracy compared with its counterparts.}
\label{tab:tab9}
\begin{small}
\begin{tabular}{lccc}
\hline
\multicolumn{1}{c}{Wide ResNet34} & Clean            & PGD-20 Attack    & \begin{tabular}[c]{@{}c@{}}Time Cost \\ (mins)\end{tabular} \\ \hline
Natural                           & \textbf{95.03\%} & 0.00\%           & 233                                                         \\ \hline
PGD-10 \cite{madry2017towards}                           & 87.30\%          & 47.04\%          & 2713                                                        \\
Free-8 \cite{shafahi2019adversarial}                           & 86.29\%          & 47.00\%          & 667                                                         \\
YOPO-5-3 \cite{zhang2019you}                         & 86.70\%          & 47.98\%          & 476                                                         \\ \hline
PDA-3-16                   & 89.12\%          & \textbf{48.14\%} & \textbf{408}                                                \\ \hline
\end{tabular}
\end{small}
\end{table}
%------------------------------------------------------------

%---------------Fig 2--------------------------------
\begin{figure}[h!]
\centering
%\vspace{-0.15in}
%\hspace{-0.2in}
\subfigure{
\includegraphics[width=0.46\linewidth]{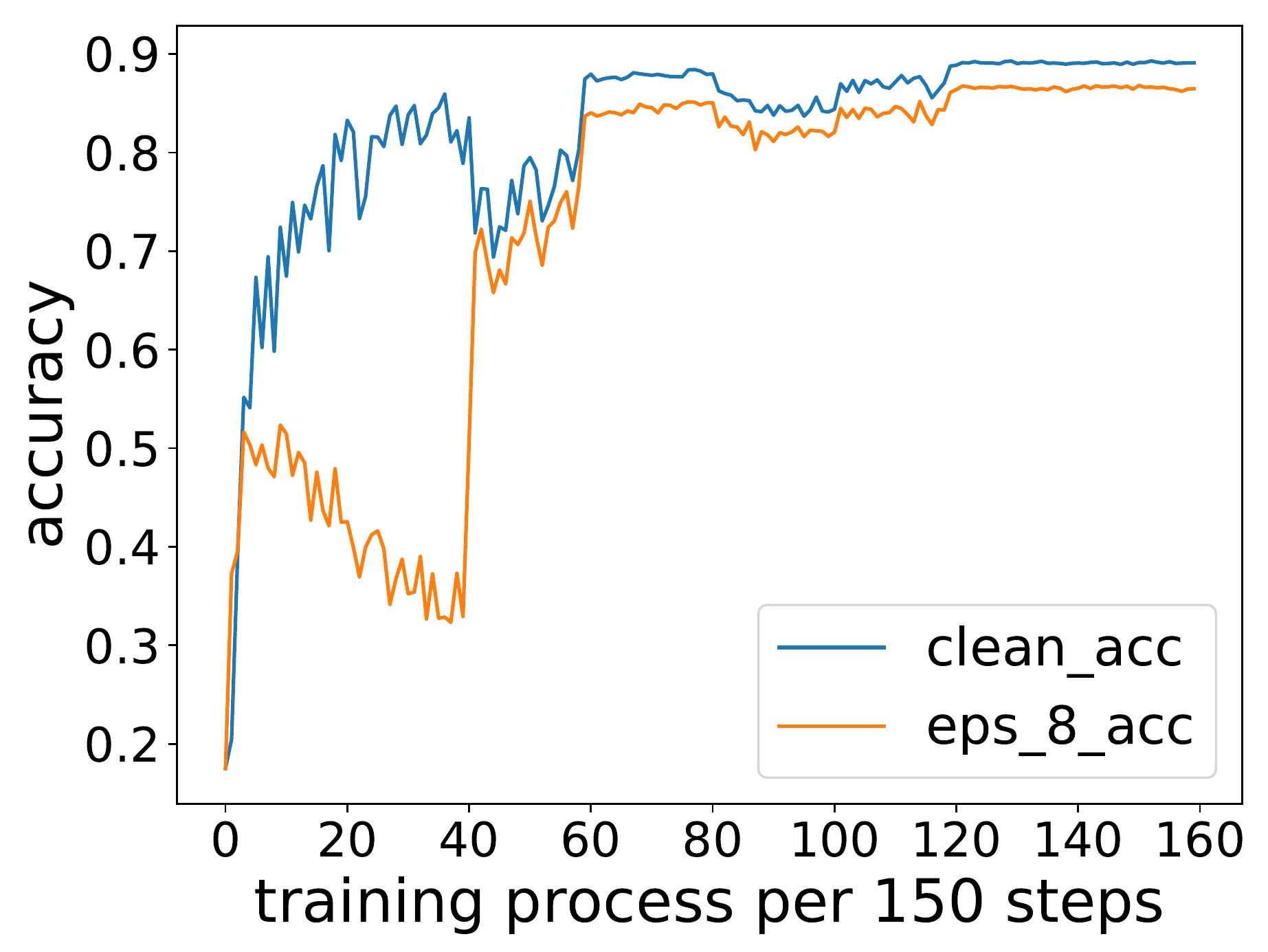}
}
%\hspace{-0.1in}
\subfigure{
\includegraphics[width=0.46\linewidth]{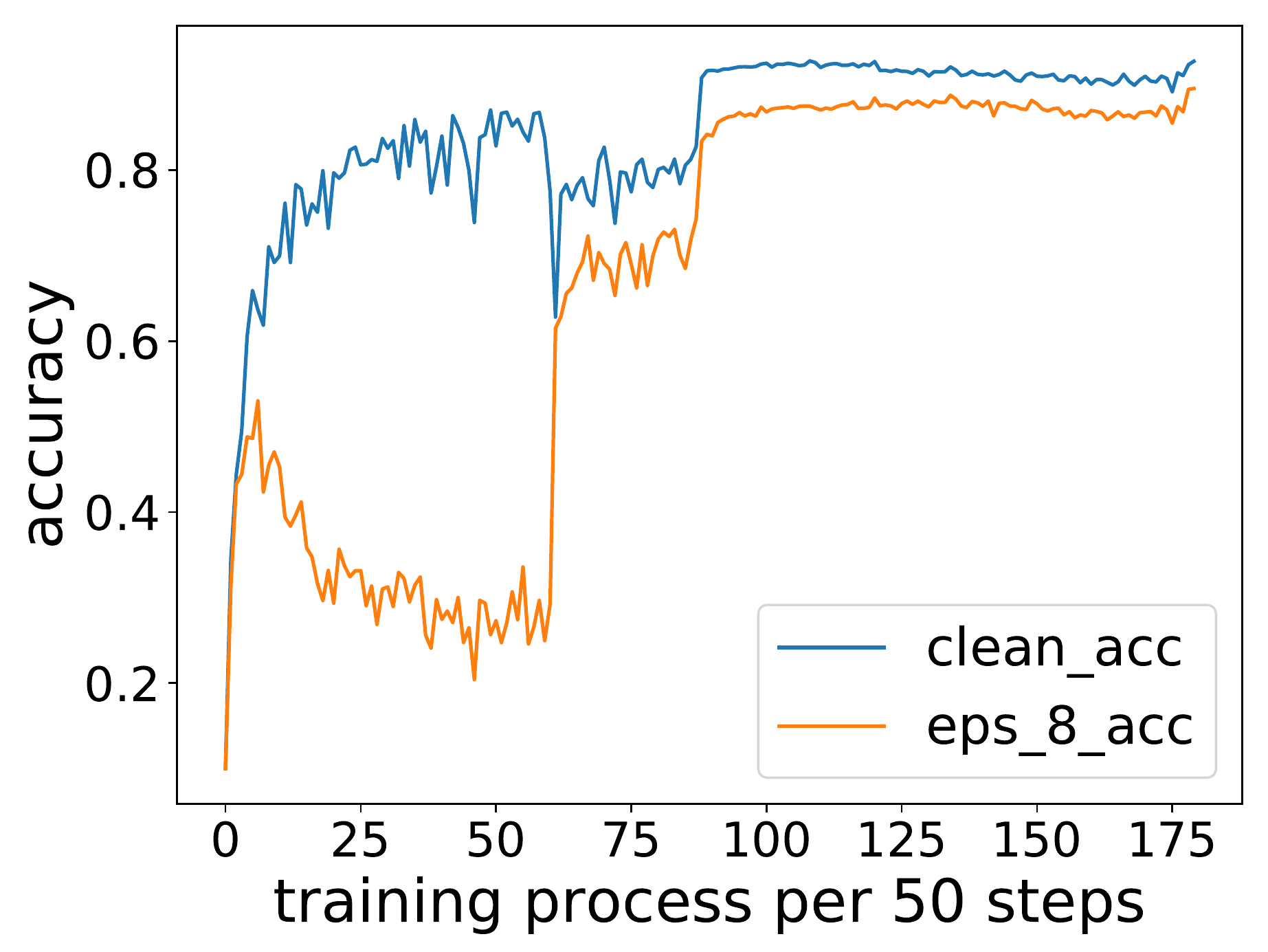}
}
%\vspace{-0.1in}
\caption{The change of accuracy in the training process of model with PDA using VGG16 (left) and Wide ResNet34 (right) on CIFAR-10. The blur curve represents clean accuracy and the coral curve denotes the performance against black-box adversarial examples by PGD-20 attack ($\varepsilon$=8/255). } %The type of examples is the same as former experiments, which contains benign examples, $l_{inf}$ norm adversarial examples from PGD, $l_{2}$ norm adversarial examples from C\&W, and corruption sets for blur, noise and other.}
\label{fig:fig2}
%\vspace{-0.2in}
\end{figure}
%-----------------------------------------------------------

In Figure \ref{fig:fig2}, we provide the accuracy curves using PDA in the training process on CIFAR-10. For the model with PDA strategy, the additive noises are diversified in many phases. In the first drop of coral curve (`eps\_8\_acc') without additive noise of PDA, we could notice that the adversarial robustness starts to decrease after a period of increase, while the clean accuracy keeps increasing. This can be attributed to the roughness of decision boundary and the tendency of overfitting. After more varied data is augmented, the black-box adversarial robustness tends to increase in accordance with the clean accuracy.

\paragraph{SVHN} We use standard ResNet18 model to evaluate the adversarial robustness on SVHN. Due to the lack of related corruption set, we only test the adversarial performance of models. Table \ref{tab:tab4} shows that our PDA achieves better performance on both adversarial attack and clean images, in comparison with other augmentation strategies.

%---------------------------------Tab 4----------------------
\begin{table}[h!]
\caption{The results of adversarial robustness for ResNet18 on SVHN. The attack step size $\alpha$ of PGD-20 attack is the same as CIFAR-10. PDA performs better than PGD and GDA against adversarial attack as well as clean images.}
\label{tab:tab4}
\begin{small}
\begin{tabular}{lcccc}
\hline
ResNet18  & Clean            & \multicolumn{2}{c}{\begin{tabular}[c]{@{}c@{}}PGD-20 Attack\\ ($\varepsilon$=4, 8)\end{tabular}} & \begin{tabular}[c]{@{}c@{}}CW Attack\\ (c=500)\end{tabular} \\ \hline
Natural   & \textbf{96.42\%} & 43.51\%                                        & 10.37\%                                        & 8.89\%                                                      \\
PGD-10-1 \cite{madry2017towards} & 93.23\%          & 64.56\%                                        & 38.30\%                                        & 19.70\%                                                     \\
GDA-0.1 \cite{ford2019adversarial}  & 94.72\%          & 59.62\%                                        & 30.74\%                                        & 16.11\%                                                     \\ \hline
PDA-3-1.0 & 96.11\%          & \textbf{72.34\%}                                        & \textbf{47.08\%}                                        & \textbf{21.58\%}                                            \\ \hline
\end{tabular}
\end{small}
\end{table}
%--------------------------------------------------------------

\paragraph{ImageNet} Due to the limited space of device, we randomly select 200 out of the 1000 classes of original ImageNet, and restrict the training set as 100k and the validation set as 10k. We use AlexNet to evaluate the adversarial robustness against PGD-20 attack and BPDA attack (ICML 2018 best paper). The results in Table \ref{tab:tab6} demonstrate that PDA has the best clean accuracy outperforming the other methods. Besides, it can be observed that Gaussian noise augmentation has little improvement on adversarial robustness as the image sizes increase on ImageNet.

%For GDA at ImageNet scale, the positive relation between adversarial robustness and Gaussian noise is much weaker compared with CIFAR-10.
%---------------------------------Tab 6----------------------
\begin{table}[h!]
\caption{The results of adversarial robustness for AlexNet on ImageNet. PDA has comparable adversarial robustness and better clean accuracy than its counterparts.}
\label{tab:tab6}
\begin{small}
\begin{tabular}{lccc}
\hline
AlexNet  & Clean            & \begin{tabular}[c]{@{}c@{}}PGD-20 Attack\\ ($\varepsilon$=8)\end{tabular} & \begin{tabular}[c]{@{}c@{}}BPDA Attack\\ ($\varepsilon$=8)\end{tabular} \\ \hline
Natural  & \textbf{59.08\%} & 2.40\%                                                                    & 7.90\%                                                                  \\
PGD-10-2 \cite{madry2017towards}   & 53.30\%          & 25.20\%                                                                   & 26.80\%                                                                 \\
GDA-0.4 \cite{ford2019adversarial}  & 43.42\%          & 1.67\%                                                                     & 1.02\%                                                                  \\ \hline
PDA-3-10 & 56.76\%          & \textbf{27.40\%}                                                          & \textbf{28.20\%}                                                        \\ \hline
\end{tabular}
\end{small}
\end{table}
%--------------------------------------------------------------

%Figure \ref{fig:fig2} shows the accuracy performance of different methods, from which we can get the following observations: (1) It is obvious that GDA achieves good performance for all types of corruptions, which meets part of the assumption in Theorem 1. However, the $\mathcal{R}_{adv}$ for GDA could be less than PDA due to the lack of adversarial robust augmentation, and thus leads to weak adversarial robustness. (2) PGD-AT performs poorly on corruptions and clean examples, which might derive from the decrease of $\mathcal{C}_{i}^{T}$ in Theorem 1, leading to the looser upper bound and weaker corruption robustness compared to PDA. (3) For all cases, models trained by PDA consistently achieve the most robust performance under both adversarial examples and corruptions, which means that PDA brings more data with higher complexity. In this way, $\mathcal{R}_{adv}$ shrinks significantly and $(\mathcal{C}_{i}^{T})^{\varepsilon} \backslash \mathcal{C}_{i}^{T}$ increases, which leads to a tighter upper bound and the stronger adversarial robustness as well as corruption robustness.

%---------------------------------------------------------------------
%---------------------------------Tab 2----------------------
\begin{table*}[tp!]
\caption{The corruption robustness of naturally trained model and models augmented with different strategies. The table above shows the results of VGG16 on CIFAR-10-C, and the table below represents the results of AlexNet on ImageNet-C. The detailed results are provided in the appendix. PDA has the lowest mCE and corruption errors as well as better clean error than other augmentation strategies.}
\label{tab:tab2}
\begin{center}
\begin{small}
\begin{tabular}{lccccccc}
\hline
VGG16  & Clean Error     & mCE              & RmCE             & Noise            & Blur             & Weather          & Digital          \\ \hline
Naturally Trained   & \textbf{7.48\%} & 100.00\%         & 100.00\%         & 46.22\%          & 30.34\%          & 17.99\%          & 23.64\%          \\
PGD-5-1 \cite{madry2017towards}            & 12.40\%         & 77.17\%          & 75.57\%          & 18.11\%          & 19.93\%          & 22.67\%          & 25.33\%          \\
PGD-10-1 \cite{madry2017towards}           & 16.23\%         & 88.29\%          & 85.97\%          & 22.00\%          & 22.02\%          & 26.16\%          & 28.58\%          \\
GDA-0.1 \cite{ford2019adversarial}            & 10.86\%         & 76.92\%          & 75.69\%          & 16.39\%        & 25.80\%          & 20.43\%          & 24.97\%          \\
GDA-0.4 \cite{ford2019adversarial}            & 31.45\%         & 151.32\%         & 146.47\%         & 25.86\%          & 49.01\%          & 43.86\%          & 47.57\%          \\ \hline
PDA-3-1.0           & 9.06\%          & \textbf{66.55\%} & \textbf{65.56\%} & 18.88\%          & \textbf{18.05\%} & \textbf{17.90\%} & \textbf{20.18\%} \\
PDA-3-2.0           & 10.44\%         & 70.35\%          & 69.10\%          & \textbf{15.99\%}          & 18.85\%          & 20.77\%          & 21.95\%          \\ \hline
\end{tabular}
\begin{tabular}{lccccccc}
\hline
AlexNet           & Clean Error      & mCE              & RmCE             & Noise   & Blur             & Weather          & Digital          \\ \hline
Naturally Trained & \textbf{40.92\%} & 100.00\%         & 100.00\%         & 81.89\% & 66.95\%          & 71.21\%          & 57.48\%          \\
PGD-10-2 \cite{madry2017towards}            & 56.58\%          & 107.02\%         & 101.62\%         & 86.38\% & 73.32\%          & 74.59\%          & 62.73\%          \\
GDA-0.4 \cite{ford2019adversarial}          & 46.70\%          & 117.55\%         & 118.15\%         & \textbf{71.26\%} & 92.62\%          & 84.26\%          & 78.10\%          \\ \hline
PDA-3-10 & 43.24\%          & \textbf{98.86\%} & \textbf{97.68\%} & 80.64\% & \textbf{65.99\%} & \textbf{70.76\%} & \textbf{56.98\%} \\ \hline
\end{tabular}
\vspace{-0.1in}
\end{small}
\end{center}
\end{table*}
%------------------------------------------------------------

%---------------------------------Tab 7----------------------
%\begin{table*}[tp!]
%\caption{The corruption robustness of AlexNet on ImageNet-C, the higher clean accuracy and lower mCE of PDA indicate the better general robustness.}
%\label{tab:tab7}
%\begin{center}
%\begin{small}
%\begin{tabular}{lccccccc}
%\hline
%AlexNet           & Clean Error      & mCE              & RmCE             & Noise   & Blur             & Weather          & Digital          \\ \hline
%Naturally Trained & \textbf{40.92\%} & 100.00\%         & 100.00\%         & 81.89\% & 66.95\%          & 71.21\%          & 57.48\%          \\
%PGD-10-2 \cite{madry2017towards}            & 56.58\%          & 107.02\%         & 101.62\%         & 86.38\% & 73.32\%          & 74.59\%          & 62.73\%          \\
%GDA-0.4 \cite{ford2019adversarial}          & 46.70\%          & 117.55\%         & 118.15\%         & \textbf{71.26\%} & 92.62\%          & 84.26\%          & 78.10\%          \\ \hline
%PDA-3-10 & 43.24\%          & \textbf{98.86\%} & \textbf{97.68\%} & 80.64\% & \textbf{65.99\%} & \textbf{70.76\%} & \textbf{56.98\%} \\ \hline
%\end{tabular}
%\end{small}
%\end{center}
%\end{table*}
%--------------------------------------------------------------

\subsection{Corruption Robustness}
\begin{figure}[h!]
\centering
%\vspace{-0.15in}
%\hspace{-0.2in}
\subfigure[CIFAR-10-C]{
\includegraphics[width=0.46\linewidth]{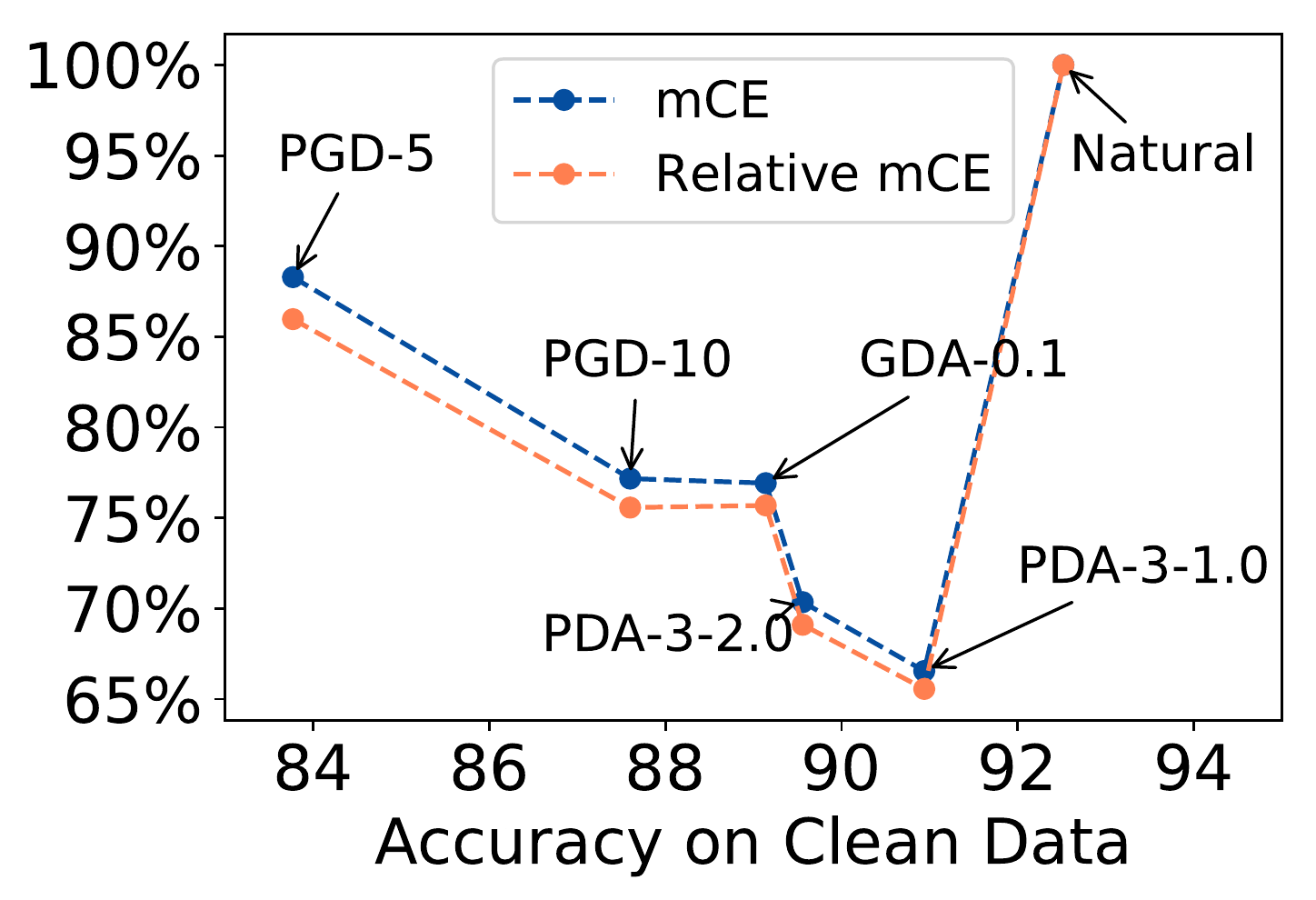}
}
%\hspace{-0.1in}
\subfigure[ImageNet-10-C]{
\includegraphics[width=0.46\linewidth]{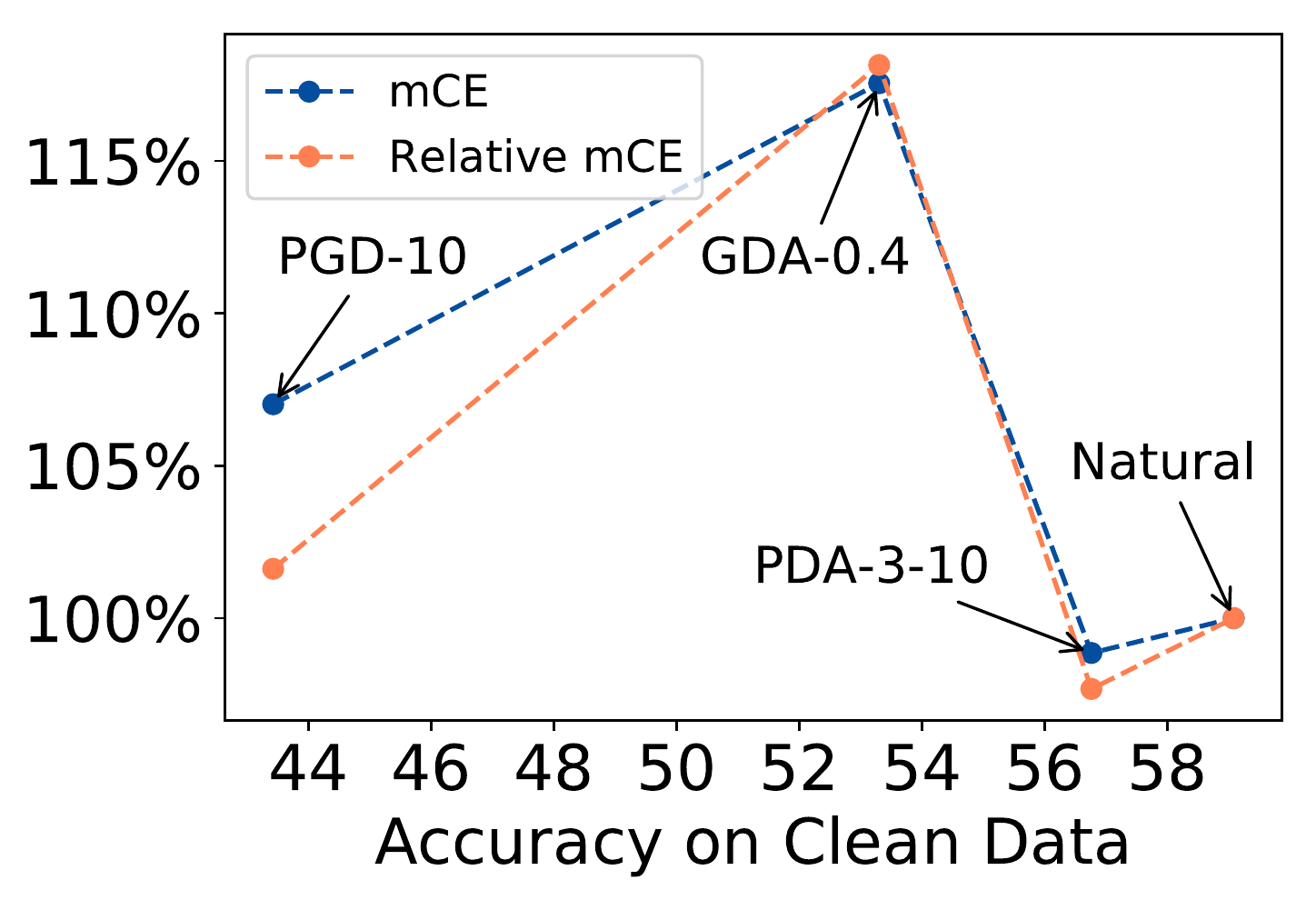}
}
%\vspace{-0.1in}
\caption{The mCE and Relative mCE of VGG16 on CIFAR-10-C and AlexNet on ImageNet-C. The higher clean accuracy and lower mCE (more lower right) indicate the better general robustness.} %The type of examples is the same as former experiments, which contains benign examples, $l_{inf}$ norm adversarial examples from PGD, $l_{2}$ norm adversarial examples from C\&W, and corruption sets for blur, noise and other.}
\label{fig:fig3}
%\vspace{-0.2in}
\end{figure}

In addition to the adversarial examples of worst-case bounded perturbations, we further test more general sets of corruptions that are likely to encounter in real-world settings. Similar to the benchmark \cite{hendrycks2018benchmarking}, we divide all 15 corruptions into 4 sets: \textit{Noise}, \textit{Blur}, \textit{Weather} and \textit{Digital}. Table \ref{tab:tab2} compares the performances of our proposed PDA and other methods on CIFAR-10-C and ImageNet-C. On CIFAR-10-C, PDA significantly surpasses other strategies on corruption robustness, despite that the naturally trained model performs well on \textit{Digital}. On ImageNet-C, PDA performs better in most corruptions but \textit{Noise}.
%The excellent performance of GDA on \textit{Noise} can be attributed to the fact that GDA itself uses Gaussian Noise for augmentation, despite  that the error rate of GDA-0.4 has reached 69.50\% on Gaussian Noise.
Figure \ref{fig:fig3} shows the mean corruption error (mCE) and Relative mCE of models trained with different strategies on CIFAR-10-C and ImageNet-C. It can be seen that PDA has generated more varied data with better general robustness.

%{\rcheng{The following illustration should be in the figure caption}}

%Comparative methods, especially PAT, though perform good on adversarial examples, show weak robustness to both static and dynamic corruptions. Also, an interesting phenomenon can be observed that all comparative strategies even perform worse than Vanilla model for dynamic corruption (showing in higher mFR values). Most adversarial training methods attempt to inject noises to inputs by searching for the worst-case perturbations which indeed improve adversarial model robustness. However, their tactics seem to be worthless to average-case or general perturbations, and may be somehow counteractive to corruption robustness. Since robustness requires high data complexity \cite{schmidt2018adversarially}, our ANP introduces adversarial noises with high complexity and diversity via progressive iteration contributing both adversarial and corruption robustness. Thus, we can draw the conclusions that ANP supplies models with strong corruption robustness compared to other defense methods.

%---------------Fig 3--------------------------------
%-----------------------------------------------------------

\subsection{Robustness in the Frequency-based analysis}

\begin{figure}[h!]
\centering
%\vspace{-0.15in}
%\hspace{-0.2in}
%\hspace{-0.1in}

\includegraphics[width=\linewidth]{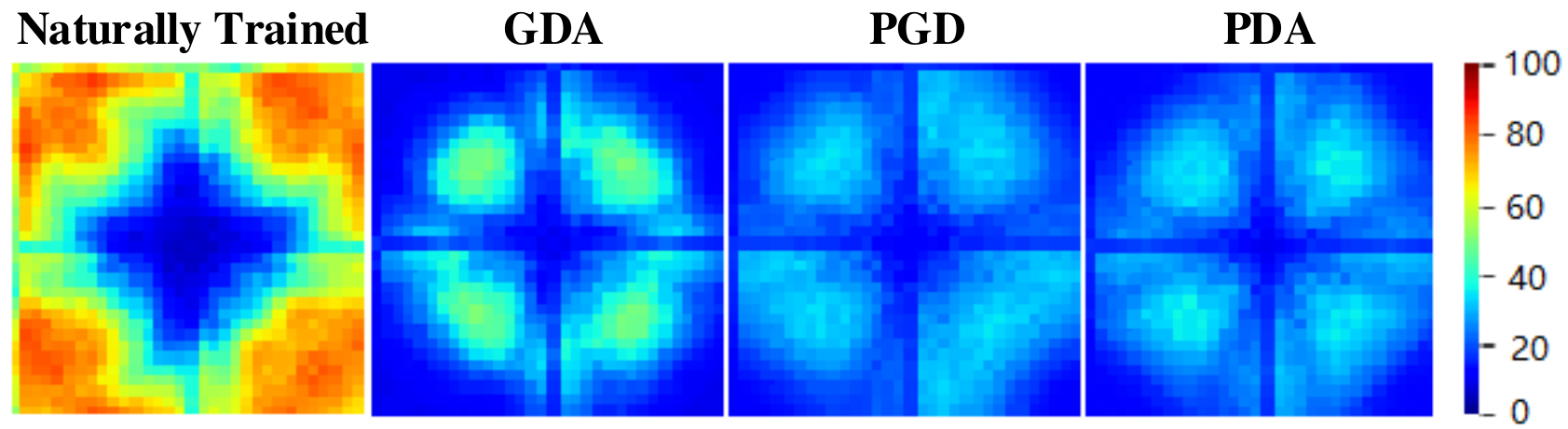}
%\vspace{-0.1in}
\caption{The heat map of model test errors with images perturbed by Fourier basis vectors using different strategies on CIFAR-10.} %The type of examples is the same as former experiments, which contains benign examples, $l_{inf}$ norm adversarial examples from PGD, $l_{2}$ norm adversarial examples from C\&W, and corruption sets for blur, noise and other.}
\label{fig:fig4}
%\vspace{-0.1in}
\end{figure}

\begin{figure}[h!]
\centering
%\vspace{-0.15in}
%\hspace{-0.2in}
%\hspace{-0.1in}

\includegraphics[width=\linewidth]{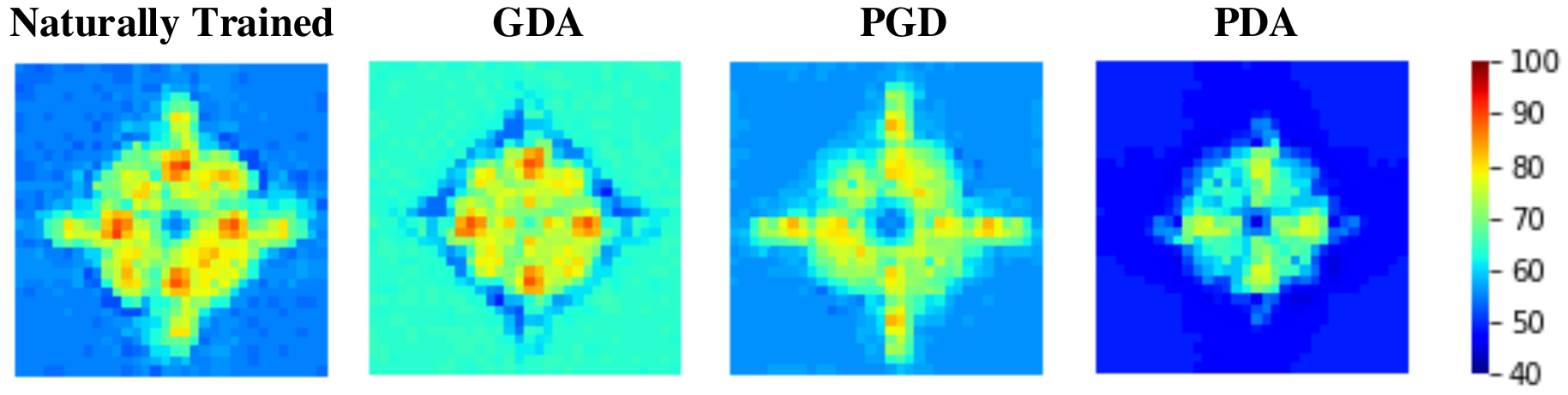}
%\vspace{-0.1in}
\caption{The heat map of model test errors with images perturbed by Fourier basis vectors using different strategies on ImageNet.} %The type of examples is the same as former experiments, which contains benign examples, $l_{inf}$ norm adversarial examples from PGD, $l_{2}$ norm adversarial examples from C\&W, and corruption sets for blur, noise and other.}
\label{fig:fig7}
\vspace{-0.1in}
\end{figure}

To better understand the performance of PDA, we conduct the experiment with frequency-based analysis  using the method proposed in \cite{yin2019fourier}. First, we perturb each image with noise sampled at each direction and frequency in Fourier domain. Then, we evaluate natural and augmented models with the Fourier-noise-corrupted images and obtain the test errors. Finally, we present the changes of error rates with heat map in Figure \ref{fig:fig4} and \ref{fig:fig7}, both indicating the model sensitivity to different frequencies and orientation perturbations in the Fourier space. For CIFAR-10 and ImageNet, we add noise with 10\% of each image's norm.

%---------------Fig 4--------------------------------
%-----------------------------------------------------------

\begin{figure}[h!]
\centering
%\vspace{-0.15in}
%\hspace{-0.2in}
%\hspace{-0.1in}
\subfigure[CIFAR-10]{
\includegraphics[width=0.3\linewidth]{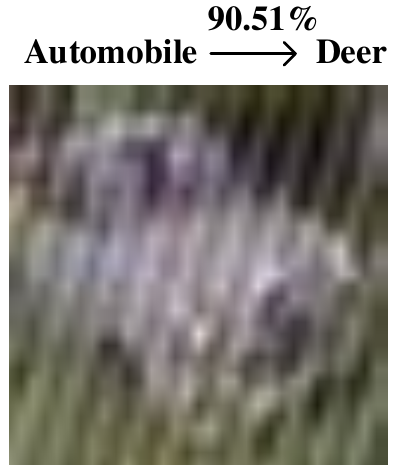}
}
\hspace{0.35in}
\subfigure[ImageNet]{
\includegraphics[width=0.3\linewidth]{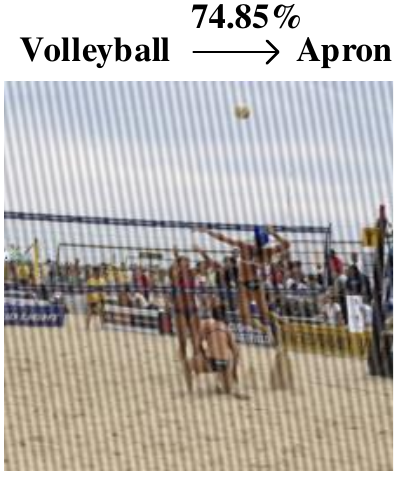}
}
%\vspace{-0.1in}
\caption{Example images perturbed by Fourier basis vectors with large norm, which misleads model successfully and preserves the content.} %The type of examples is the same as former experiments, which contains benign examples, $l_{inf}$ norm adversarial examples from PGD, $l_{2}$ norm adversarial examples from C\&W, and corruption sets for blur, noise and other.}
\label{fig:fig5}
%\vspace{-0.2in}
\end{figure}

The results show different sensitivities for naturally trained and augmented models. On CIFAR-10, the naturally trained model is sensitive to the high frequencies, where augmentation methods improve robustness with less sacrifice of lower frequencies. We notice that PDA and PGD are less sensitive in lower frequency area than GDA, and PDA appears evenly robust due to the varied data in augmentation. On ImageNet, GDA has little improvement on robustness to higher frequencies (which may need stronger additive noise); in contrast, the PDA-based model improves robustness for both lower and higher frequency noise. Figure \ref{fig:fig5} shows that,  in some cases,  adding Fourier-noise with large norm on images is similar to the black-box adversarial attacks, which can preserve the content but mislead the models.

%The naturally trained model is highly sensitive to additive perturbations in all but the lowest frequencies, while GDA and PGD both dramatically improve robustness in the higher frequencies. For the models trained with data augmentation, we see a subtle but distinct lack of robustness at the lowest frequencies (relative to the naturally trained model). Figure 4 shows similar results for three different models on ImageNet. Similar to CIFAR-10, Gaussian data augmentation improves robustness to high frequency perturbations while reducing performance on low frequency perturbations
%First, we test model sensitivity to perturbations along each Fourier basis vector.
%For CIFAR-10 models, we present this analysis for the entire Fourier domain, with noise sampled with norm 4. For ImageNet, we focus our analysis on lower frequencies that are more visually salient add noise with norm 15.7.

%---------------Fig 4--------------------------------
%-----------------------------------------------------------

\subsection{Mixed Test for Generalization Ability}
As indicated in \cite{xu2012robustness}, if the test data is similar to training data, then the corresponding test error should be close to the empirical training error. Previous adversarial related studies \cite{athalye2018obfuscated,xie2018mitigating} usually evaluate the accuracy of model on individual type of examples. However, for models trained on dataset augmented with noises, it can be prejudiced to evaluate model generalization ability on the specified type of test data different from the training set. Therefore, it is reasonable to evaluate model generalization ability with mixed types of samples when the model is trained with data augmentation containing noises. Thus, we propose \emph{Mixed Test} to fairly evaluate the model generalization ability of the proposed PDA method. Specifically, the test data combines clean, adversarial and corrupted examples in an equal proportion.
%---------------Fig 5--------------------------------
\vspace{-0.1in}
%-----------------------------------------------------------

The experimental results using \emph{Mixed Test} on CIFAR-10 are shown in Table \ref{tab:tab3}, where `C' denotes clean images, $\varepsilon$4 means the adversarial examples, \textit{Other} = \{\textit{Weather}, \textit{Digital}\} represents the corruptions mentioned above. See results of SVHN in the appendix. As can be observed, PDA outperforms the other data augmentation methods in terms of the generalization ability. As illustrated in Figure \ref{fig:fig1}, although PGD and GDA increase model robustness against specific noise, they miss some portions of benign examples, thus leading to weaker generalization ability.
%---------------------------------Tab 3----------------------
\begin{table}[h!]
\caption{The results of mixed test with VGG16 on CIFAR-10. PDA performs comprehensively well on mixed test, which confirms the superior general robustness of PDA.}
\label{tab:tab3}
\begin{small}
\begin{tabular}{lccc}
\hline
VGG16             & \multicolumn{1}{c}{C+$\varepsilon$4+Blur} & \multicolumn{1}{c}{C+$\varepsilon$4+Noise} & \multicolumn{1}{c}{C+$\varepsilon$4+Other} \\ \hline
Natural           & 60.53\%                       & 57.27\%                        & 63.69\%                        \\
PGD-10-1 \cite{madry2017towards}         & 75.40\%                       & 75.71\%                        & 75.69\%                        \\
GDA-0.1 \cite{ford2019adversarial}          & 71.00\%                       & 72.95\%                        & 72.87\%                        \\ \hline
PDA-3-1.0         & \textbf{79.75\%}              & \textbf{80.43\%}               & \textbf{80.48\%}               \\ \hline
\end{tabular}
\end{small}
\end{table}
%-------------------------------------------------------------

%The results on SVHN are provided in appendix.
%Also, it is interesting to see that the accuracy of GDA and ERM (Naive) drops more drastically compared to PGD-AT and PDA, when adversarial perturbation (e.g. $\varepsilon$ or $c$) increases. This phenomenon further demonstrates the strong adversarial defense ability of PGD-AT and PDA.

\subsection{Effects of Progressive Iterations}
\begin{figure}[h!]
\centering
%\vspace{-0.15in}
%\hspace{-0.2in}
\subfigure{
\includegraphics[width=0.46\linewidth]{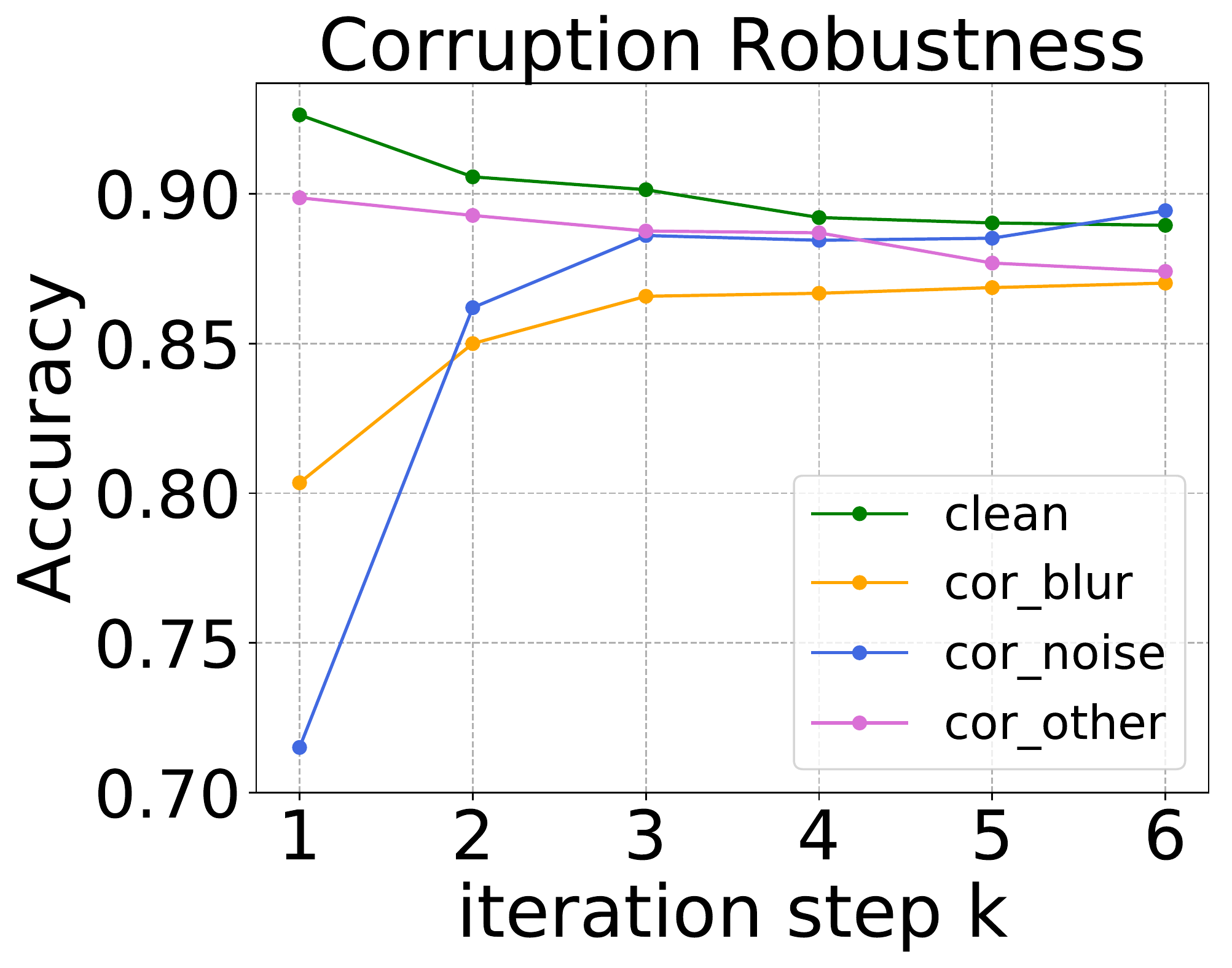}
}
%\hspace{-0.1in}
\subfigure{
\includegraphics[width=0.45\linewidth]{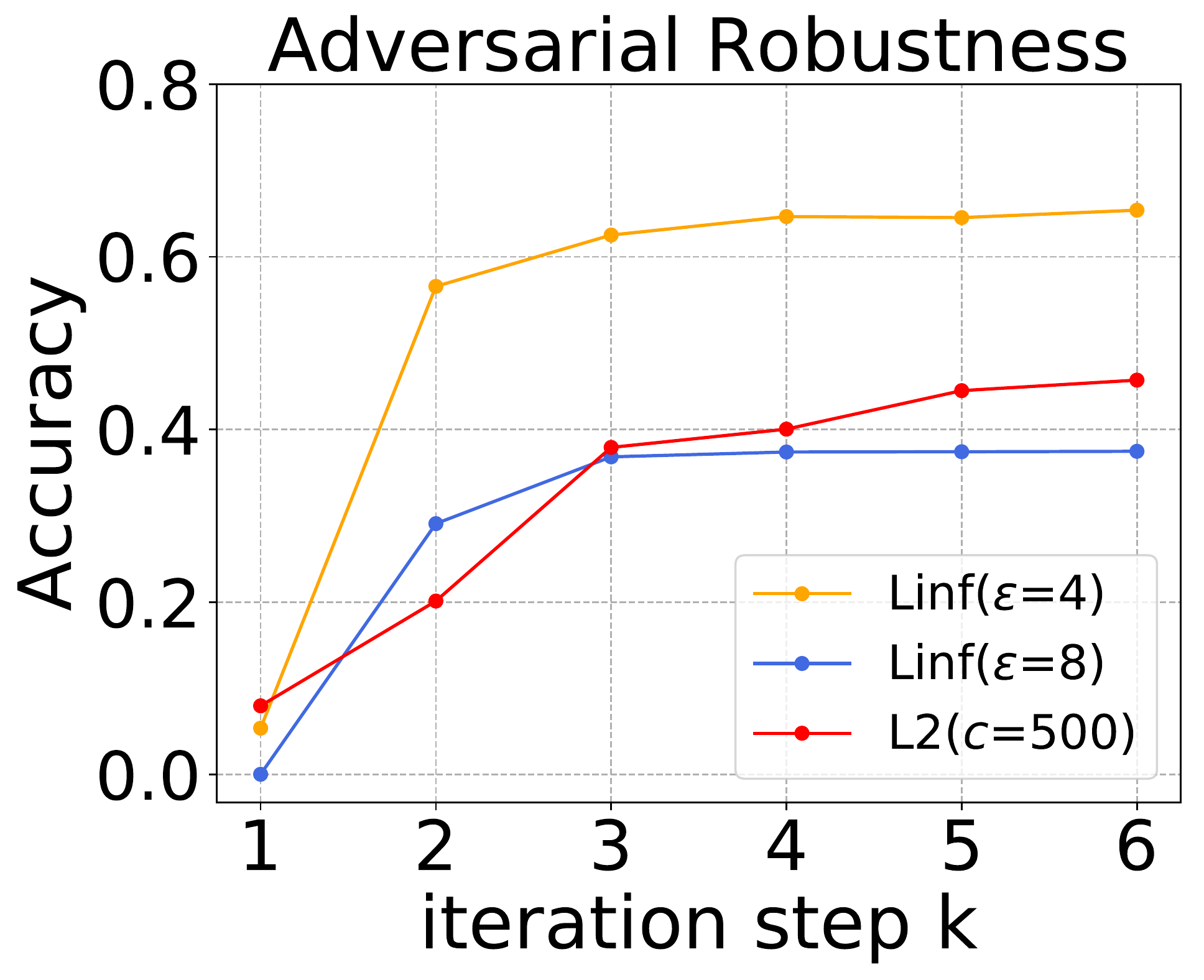}
}
%\vspace{-0.1in}
\caption{The corruption and adversarial robustness evaluation of PDA on CIFAR-10, where the range of k is from 1 to 6.} %The type of examples is the same as former experiments, which contains benign examples, $l_{inf}$ norm adversarial examples from PGD, $l_{2}$ norm adversarial examples from C\&W, and corruption sets for blur, noise and other.}
\label{fig:fig6}
\vspace{-0.1in}
\end{figure}

The multiple progressive iterations of PDA help improve the noise diversity for augmented models. To further investigate the effect of progressive iterations, we conduct an experiment on PDA with different iterative step $k$ from 1 to 6. As shown in Figure \ref{fig:fig6}, the model robustness drops as the clean accuracy increases, indicating a trade-off between accuracy and robustness. Despite that the larger $k$ brings more data complexities in the augmentation as $\varepsilon$ is fixed, it leads to more training time and less improvement on robustness. Therefore, in practice, we set $k$=3 in order to keep the balance between robustness, accuracy and computation cost.
%---------------Fig 6--------------------------------
%-----------------------------------------------------------

\subsection{Feature Alignment with Human Vision}
\begin{figure}[h!]
\centering
\includegraphics[width=0.8\linewidth]{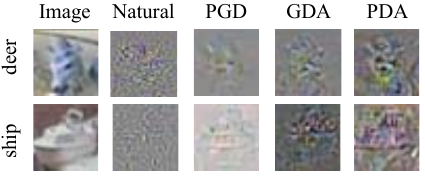}
\caption{Visualization of the gradients with respect to the input images via different training strategies on CIFAR-10.}
\label{fig:3}
\vspace{-0.1in}
\end{figure}

Beyond measuring the performance with quantitative results, we also try to discover model robustness more perceptually with visualization techniques. Since the
%\rcheng{what do you mean by 'stronger'?}
model with stronger adversarial robustness generates gradients that are better aligned with human visual perception \cite{tsipras2018robustness}, we normalize and visualize the gradient of loss w.r.t the input images on CIFAR-10. In Figure \ref{fig:3}, the gradients for `Natural' model are noisy and thus appear meaningless to human, while those for PGD and GDA are less perceptually aligned with human vision such as fewer outlines. In contrast, PDA aligns the best to human visual perception.

%This is mainly attributed to the fact that robustness is more likely a reliable property for models which could build more human-aligned applications.
%===================Fig4================================
%=======================================================

\section{Conclusions}
%In this paper, we devises a more general augmentation named \emph{Progressive Data Augmentation} (PDA) to improve both adversarial and corruption robustness with less hurt on clean accuracy. Different from the conventional methods that achieve model robustness by employing more training data \cite{schmidt2018adversarially,sun2019towards}, in PDA the diversified adversarial noises are aggregated, augmented and injected progressively, which are proven to be beneficial on improving both the adversarial and corruption robustness. We also theoretically prove that PDA is able to satisfy the perturbation bound and promise better generalization ability. Extensive experiments are conducted on datasets including CIFAR-10, SVHN, ImageNet, CIFAR-10-C and ImageNet-C which further indicate that PDA shows comprehensively excellent results on both adversarial and corruption noise compared to various augmentation methods.

In this paper, we have proposed a more general augmentation method, \ie the Progressive Data Augmentation (PDA),  which adds diversified adversarial noises progressively during training and intends to obtain more general robustness against both adversarial attacks and corruptions. Moreover, PDA has less negative impacts on clean accuracy and cheaper computation cost. We have also theoretically proved the perturbation bound and the generalization ability of PDA. With the frequency-based analysis, we have found that the model trained with PDA is more evenly robust to all kinds of frequencies. We have also noticed that the model trained with PDA has better generalization ability as evaluated more fairly on the mixed dataset. Experimental evaluations on CIFAR-10, SVHN, ImageNet have demonstrated that PDA performs comprehensively well on adversarial attacks and clean images, having achieved state-of-the-art corruption robustness on the CIFAR-10-C and ImageNet-C benchmarks.
%-------------------------------------------------------------------------

{\small
\bibliographystyle{ieee_fullname}
\bibliography{references}
}

\end{document}